# DT-Net: A novel network based on multi-directional integrated convolution and threshold convolution

Hongfeng You, Long Yu, Shengwei Tian, Xiang Ma, Yan Xing and Xiaojie Ma

**Abstract**—Since medical image data sets contain few samples and singular features, lesions are viewed as highly similar to other tissues. The traditional neural network has a limited ability to learn features. Even if a host of feature maps is expanded to obtain more semantic information, the accuracy of segmenting the final medical image is slightly improved, and the features are excessively redundant. To solve the above problems, in this paper, we propose a novel end-to-end semantic segmentation algorithm, DT-Net, and use two new convolution strategies to better achieve end-to-end semantic segmentation of medical images. 1. In the feature mining and feature fusion stage, we construct a multi-directional integrated convolution (MDIC). The core idea is to use the multi-scale convolution to enhance the local multi-directional feature maps to generate enhanced feature maps and to mine the generated features that contain more semantics without increasing the number of feature maps. 2. We also aim to further excavate and retain more meaningful deep features reduce a host of noise features in the training process. Therefore, we propose a convolution thresholding strategy. The central idea is to set a threshold to eliminate a large number of redundant features and reduce computational complexity. Through the two strategies proposed above, the algorithm proposed in this paper produces state-of-the-art results on two public medical image datasets. We prove in detail that our proposed strategy plays an important role in feature mining and eliminating redundant features. Compared with the existing semantic segmentation algorithms, our proposed algorithm has better robustness.

——————————— ◆ ———————————

## 1 INTRODUCTION

CONVOLUTIONAL neural networks (CNNs) play an active role in various visual tasks, including target detection [1] [2]; semantic segmentation [3] [4] [5] [48]; image repair [6]; and image generation [7] [8]. In recent years, CNNs have also obtained many application results in medical image segmentation. As medical images have become available in multiple forms and multiple modes, the model structure of CNNs has been continuously improving to adapt to increasingly complex medical image data sets [9] [10] [11] [12] [13]. The characteristics of medical images include small data set sample sizes, uneven target distributions, small differences between foregrounds and backgrounds, etc. Therefore, to better complete various tasks related to medical images, researchers often implement images before model training enhanced processing. Yanzhe X. et al. [14] first smoothed medical images by using the difference of Gaussian (DoG) filter, then analysed and identified candidate spots according to the Hessian filter, and next combined them with the U-Net to capture and identify the most likely spot location. Finally, working with two public medical image data sets, they proved the feasibility of the algorithm. Soomro T. A. et al. [15] first removed uneven illumination through fuzzy calculations and morphological operations, thereby selecting RGB channels with better contrast. Then, the improved U-Net model was used to achieve more accurate image segmentation and accelerate the entire training process. The experimental results obtained higher sensitivity and accuracy on the two public data sets of DRIVE and STARE. To better analyse cancers and brain tumours in medical images, Patel S. et al. [17] introduced a histogram equalization technology to enhance the contrast and brightness of the image without losing the original information. To improve the diagnostic accuracy of images and reduce the impact of speckle noise, Singh P. et al. [18] proposed an adaptive histogram equalization method through the specific analysis of breast cancer, uterine fibroids, scrotal cysts, pleural effusion and other diseases. Their approach improved the textural features, contrast, decomposable details and sensitive image structure of the human visual system in the image.

his research is partially supported by Xinjiang Uygur Autonomous Region Natural Science Fund Project (2016D01C050), and Xinjiang Autonomous Region Science and Technology Talent Training Project (QN 2016YX0051). We would also like to thank our tutor for the careful guidance and all the participants for their insightful comments.(Corresponding author: L. Yu and Shengwei Tian)

The authors are with Xinjiang University, 830000 Urumqi.

H. F. You is with the School of Information Science and Engineering, Xinjiang University, 830000, China (e-mail: 1053109177@qq.com).

S. W. Tian is with the Software College, Xinjiang University, 830000, China (e-mail: tianshengwei@163.com).

L. Yu is with the Network Center, Xinjiang University, 830000, China (e-mail: yul_xju@163.com).

X. Ma is with Department of Cardiology, The First Affiliated Hospital of Xinjiang Medical University,, 830011, China (e-mail: maxiangxj @163.com).

X. J Ma is Postgraduate degree, master degree. Deputy Chief Physician. The Fourth People's Hospital of Urumqi. (e-mail:maxiaojiexj@163.com).

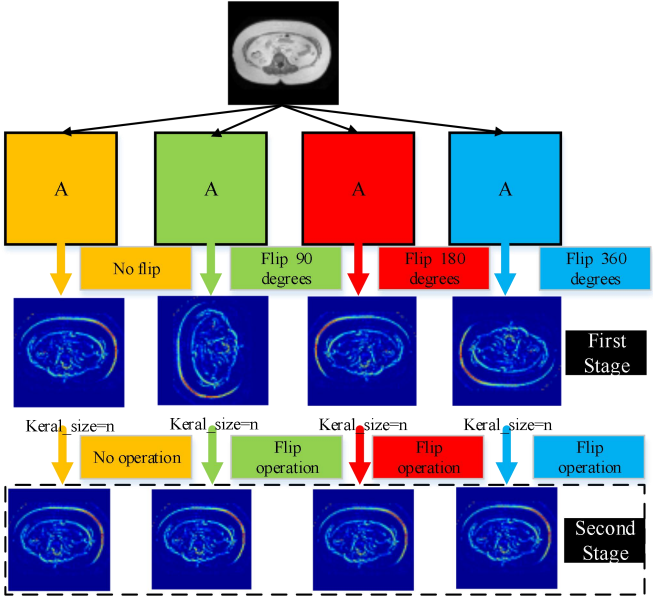

Fig. 1. Illustration of the MDIC module. The figure above only shows one quarter of the global feature maps. "A" represents a quarter of the number of feature maps; "Kernel_size" represents the size of the convolution kernel; "First Stage" represents different flip methods where different direction semantic image feature maps are generated; and "Second Stage" represents the reverse flip of the feature map after convolution in different directions.

From the above research results, it is clear that preprocessing the medical image before training can improve the quality of some features in the image and ultimately improve the recognition accuracy of the image. Most of the preprocessing is a special processing method for a certain data set to better express and emphasize the semantic information of this data set, so the effect is often not good on other datasets. Recently, the rapid development of machine learning and deep learning, especially the excellent performance of CNNs in various visual tasks, has enabled a large number of researchers to realize full mining of semantic image information features by constructing algorithms based on CNNs. These algorithms not only reduce the tedious work of preprocessing but can also be applied to a variety of image data sets and obtain good prediction results.

The Fully Convolutional Networks (FCN) [19] is the first end-to-end neural network based on CNNs, and it has led deep learning algorithms into the field of semantic image segmentation. This model overcomes the constraints of artificial features and, at the same time, makes great progress in speed and accuracy. Based on the study of FCNs, researchers found that the quality of feature maps generated by convolutional neural networks directly affects the prediction results of the final model. Gu Z. et al. [20] proposed a context encoder network that solved the problem of some spatial information loss caused by the continuous pooling of U-Net and strided convolutional operations. The main idea of the network is to use the ResNet block as a fixed-feature extractor and generate higher-level semantic feature maps through dense and residual network blocks. The feasibility of the algorithm was verified on three public medical image data sets. Weng Y. et al. proposed an NAS-U-Net algorithm, which reconstructed the encoder and decoder of U-Net by constructing an upsampling unit structure and a downsampling unit structure. Under the premise of reducing parameters, the model realized feature map refinement and finally obtained good performance in MRI, computed tomography, and ultrasound acquisition. Zhou S. et al. [22] introduced a fuzzy medical image segmentation algorithm to locate the boundaries of complex medical images and segment small targets correctly. The algorithm uses the U-Net and the HED structure to obtain a comprehensive multi-scale information feature map. At the same time, it uses multi-scale, densely linked network blocks to obtain accurate semantic feature map information and obtain fuzzy boundary positioning. A host of experiments showed that their proposed model was effective at predicting fuzzy feasibility on medical images. Castro E. et al. [16] proposed a weight rotation strategy based on convolution, which used random rotation weights to enhance the model's ability to learn from data and generate feature maps with different weights. This algorithm solves the problem where a small amount of data makes it easy to overfit the model while acquiring rich feature information.

Through the above research, it is found that different model structures based on convolutional neural networks can generate different feature mapsfeature maps (which contain different semantic information about medical images). Due to the characteristics of small medical image data sets and the increased complexity of images, more feature maps are often needed to obtain more deep semantic information. Therefore, in the task of semantic segmentation of medical images, the quality of feature maps has become particularly important. In recent years, some models [23][24][25] have used multi-scale convolutions or multiple channels [26][27] to generate more shallow feature maps and enable deep feature maps to contain richer semantic information. Whether by increasing the feature map or through the fusion of multiple feature maps, a host of calculation parameters are generated, thereby increasing the complexity of the model. This method is able to obtain higher-quality feature maps and does not increase the number of feature maps in small-sample medical data sets, so it can provide more semantic information for the final semantic segmentation of the medical image. In this paper, we construct two new feature map extraction strategies, thereby better realizing the task of small-sample medical image segmentation. The main contributions of this article are as follows.

1) We propose a novel convolution method: multi-directional integrated convolution (MDIC). The core idea of the convolution is to divide the feature map of the previous layer into four parts equivalently and flip each part to 0 degrees, 90 degrees, 180 degrees and 360 degrees. Then, each part uses convolution kernels of different sizes to obtain the semantic information about the part in different directions. In this article, we call the feature map of each part a local feature map. Then, we fuse the local feature maps from different parts that share the same direction but contain different convolutions to generate multi-directional, multi-scale feature maps that contain richer semantic information. The MDIC strategy enhances the ability of model to mine features without increasing the number of feature maps through this method of strengthening local feature maps. Figure 1 shows the MDIC model's description of a set of local feature maps.

2) The enhancement of local feature maps and the integration of global feature maps combine to produce a host of feature maps containing rich elements. To further mine and retain the deep features that contain more semantics during training, we propose a convolution threshold strategy. The core idea of this strategy is to assign a weight coefficient to all the features in each feature map through the ReLU activation function, set the threshold through the weight coefficient, and retain the features larger than the threshold while clearing the features smaller than the threshold. This strategy can retain more deep semantic features and eliminate many redundant features.

3) We visualize the multi-directional convolution strategy so that readers better understand them and obtain state-of-the-art performance on two public data sets.

The rest of this article is organized as follows: In the second section, we introduce the advantages and disadvantages of some traditional medical algorithms and the latest medical algorithms. The third section introduces the two strategies of our proposed algorithm in detail. In the fourth section, relevant experiments are carried out for the proposed algorithm, and the influence of each strategy or parameter on the model is analysed in detail. The fifth section gives our conclusions and the next steps for further research.

## 2  RELATED WORKS

The end-to-end semantic segmentation model has achieved excellent results in medical image segmentation tasks [28] [29] [30]. The convolutional neural network, as an important part of the end-to-end model, occupies an irreplaceable position in feature map generation and fusion.

### 2.1 Feature Maps

The quality of the feature map directly affects the result of semantic image segmentation. As the first end-to-end network to realize semantic image segmentation algorithm for the first time, the FCN realized the prediction of each pixel through the upsampling layer, thereby breaking the limitation of mapping feature maps into a feature vector, and it played a very important pioneering role for later end-to-end algorithms. FCNs use multi-layer convolutions to automatically learn features, obtains more representative deep features through different receptive fields, and finally obtains a feature map containing rich semantic information. In subsequent research, researchers used convolutional neural networks to construct various end-to-end semantic segmentation algorithms and obtained many results in medical image segmentation. U-Net, proposed by O. Ronneberger et al. [31], is an improvement of the FCN algorithm. In the encoder of U-Net, the method of generating feature maps is similar to that of the FCN, which is also realized through multi-layer convolutions. In the decoder of U-Net, the generated feature map is fused with the feature map of the encoder, thereby providing more semantic information for the next layer of sub-modules. This structure achieves better segmentation accuracy. Qayyum A. et al. [32] used the ResNet residual block composed of different-scale convolutions as the encoder and decoder of the Resnet end-to-end network to generate residual feature maps. The decoder uses the same strategy as that of U-Net to fuse the encoder's residual feature maps with the decoder's residual feature maps. By fusing feature maps to obtain more semantic information, thereby improving the accuracy of segmentation for medical images, Khened M. et al. [33] proposed a fully convolutional network based on dense links. The network is dense and fast. The last layer contains the feature maps of all the previous layers. After fusion, the convolutional neural network with a convolution kernel equal to 1 realizes the integration of the feature maps, reducing the number of calculation parameters. Simultaneously, the model also integrates the feature map of the encoder during the decoding stage and finally achieves the best results on the three public medical image datasets.

Through the above research, we find that different convolution strategies generate different feature maps, thereby obtaining different semantic image information. At the same time, more semantic features can be obtained by introducing and fusing the feature maps from the encoder into the decoder. These two methods ultimately improve the task of segmenting of medical images.

### 2.2 Fusion of Multiple Feature Maps

The fusion of multi-scale feature maps in the end-to-end semantic segmentation model mainly falls into two categories: one is the feature map fusion method of multi-scale convolution through different convolution kernels; the other is a variety of different convolution strategies. The generated feature maps are fused regardless. The purpose of fusion is to obtain more feature maps, thereby obtaining more medical semantic image information.

Research on medical images has been conducted based on the end-to-end semantic segmentation algorithm of multi-scale convolution. Hao J. et al. [34] proposed a multi-scale CNNs algorithm. The algorithm uses multi-scale inputs to overcome the limitations of specific input scales so that the generated feature maps can accommodate more neighbourhood information from different angles and ultimately improve the segmentation accuracy of the algorithm for brain tumours. Xia H. et al. [35] proposed multi-scale cavity convolution (MD-Net) to reduce the influence of the segmented background on the foreground. The network uses feature pyramids constructed by the convolution of holes of different sizes to extract semantic information from medical images, thereby generating a large number of feature map. The feasibility of the algorithm was verified on two medical public data sets. Xia C. et al. [36] developed a multi-scale segmentation network (MSSN) to segment the left ventricle. The main principle is to construct parallel convolutional layers and serial convolutional layers through different hole convolutions, generate multiple feature maps, capture the multi-scale semantic features of medical images, reconstruct any missing space through upsampling layers of pixel-level information, and finally generate a high-precision segmentation model. Through the above research, we can find that a variety of feature maps generated by multi-scale convolution can provide more semantic information, so multi-scale convolution is active and effective.

The second end-to-end model for acquiring more semantic information from medical images is often implemented through multiple convolution strategies. Jafari M. et al. [37] proposed a DRU-Net algorithm. The core idea of the model is to construct a local residual operation block through multi-layer convolution and then build a dense residual network based on the residual block. To further improve the prediction accuracy of the model, the feature map of each sub-module in the encoder and the residual feature map of each residual block are simultaneously integrated into the decoder. Xia K. et al. [38] used two different convolution strategies, SCNN and ResNet, to reduce the interference of noise with the minimized feature map, thereby obtaining better representation capabilities. Qiang Z. [39] and others fully considered the advantages of the fusion of the feature maps with the decoder and encoder of U-Net, as well as the features of the fusion of feature maps between dense network blocks, and constructed a Dense-U-Net algorithm. In this algorithm, the next dense block in the encoder obtains more semantic information about the medical image by fusing the feature maps of multiple dense blocks above it. In the decoder, the dense block obtains not only the feature map of the encoder but also the feature maps of the above multiple decoding dense blocks. Through the above fusion of feature maps, a large amount of semantic information is obtained, which ultimately improves the task of segmenting medical images.

Through the above research, we find that more semantic information can be obtained in the following ways, thereby improving the accuracy of end-to-end medical image segmentation. The specific summary is as follows: 1) In the decoding stage, more semantic information can be obtained by fusing the feature maps of the encoder, thereby improving the quality of the feature maps; 2) Whether in the encoder or the decoder, the feature maps generated by the residuals, densely generated feature maps and feature maps generated by multi-layer convolution all contain rich semantic information and yield high segmentation accuracy; 3) In the same sub-module, the fusion of multiple feature maps can also yield more semantic information.

To obtain feature maps that contain richer semantic information, in this article, we construct a new convolution strategy. We call it a multi-directional integrated convolution strategy, and its purpose is to generate feature maps that contain more semantic information. The core principle of MDIC is to divide all the feature maps of the upper layer into four equal parts and perform flip operations on these four local feature maps. After the flip, convolution operations are performed on the local feature maps of each part. Then, the local feature maps of different convolutions and different parts are merged. The final feature map generated in this way not only obtains semantic information in different directions but also semantic information of different scales, which greatly improves the quality of the feature map. In the end-to-end model we build, both the encoder and the decoder are based on MDIC to achieve feature map generation. To further improve the quality of the feature map, we introduce another set of convolutions (not divided into four equal parts),called the global feature map. The convolutional kernel size of this set of convolutions and the convolutional size of the local feature map are different. The purpose of this is to obtain additional semantic information. Although the above method generates a feature map with rich semantics, not all feature points in the feature map contain a large amount of semantic information. Some of the feature points are less relevant to the target, or the amount of redundant information is greater than the amount of useful semantic information, which ultimately leads to a decrease in algorithm accuracy. To further improve the quality of feature maps and eliminate redundant features, we propose a second new convolution strategy called threshold convolution. Its main task is to set a threshold for the feature map of each sub-module in the encoding stage to eliminate some feature points that are less connected to the target. Finally, we verify the CHAOS medical image segmentation and BraTS medical image segmentation data sets. Compared with some recent medical image semantic segmentation models, our new model obtains state-of-the-art results.

## 3   METHODOLOGY

In this section, we introduce our new end-to-end semantic segmentation model, the DT-Net algorithm, in four parts: Part A describes the overall structure of the DT-Net algorithm; Part B introduces the feature maps of the MDIC module of the encoder in the mining process; Part C introduces the processes of feature map mining and fusion by the MDIC module in the decoder; Part D introduces the principle of threshold convolution.

### 3.1 DT-Net

As shown in Figure 2, the DT-Net algorithm is mainly composed of an encoder and a decoder. In the first stage, to reduce our workload, we convert the length and width of the medical images in the data set to 256 pixels each and use them as the input of the DT-Net algorithm The second stage introduces the model structure and parameter configuration of each module of the encoder. The encoder obtains feature maps with more semantics through MDIC and threshold convolution and simultaneously eliminates a host of redundant features in the feature maps. The third stage introduces the model structure and parameter configuration of each module in the decoder. The decoder obtains high-quality feature maps through MDIC and simultaneously realizes residual operations with the feature maps in the encoder to obtain more semantic information. Finally, the classifier of DT-Net algorithm for semantic segmentation is introduced.

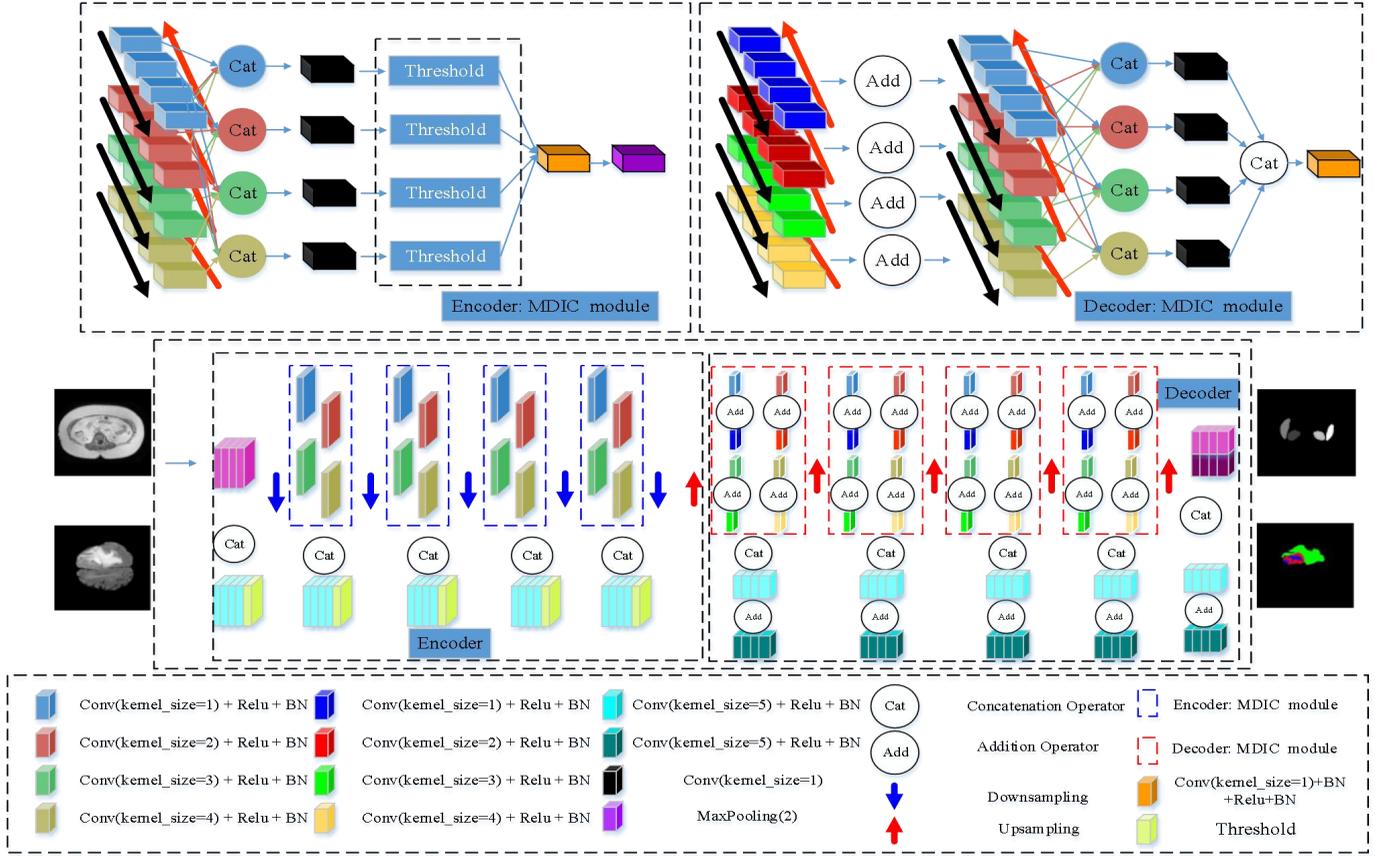

Fig. 1. Illustration of the DT-Net model. Both the encoder and decoder are composed of 5 MDIC modules. All convolutional layers realize the direct splicing of different convolution kernels through zero-padded operations. The number of fitters in sub-modules 1-5 in the encoder are [24, 48, 96, 192, 192]. The number of fitters in sub-modules 1-5 in the decoder are [192, 192, 96, 48, 24]. "Threshold" represents the threshold algorithm after convolution, and a detailed explanation is given in the section on the threshold convolution module. The black direction symbols in "Encoder: MDIC module" and "Decoder: MDIC module" represent the flip angle of each part of the feature map [0, 90, 180, 360]; the red direction symbols represent the reverse flip angle of each part [0, 90, 180, 360]. The blue and light blue matrices represent the first part of the feature map; red and light red represent the second part of the feature map; green and light green represent the third part of the feature map; yellow and light yellow represent the fourth part of the feature map.

### 3.2 Encoder: MDIC Module

The encoder of the DT-Net network is composed of 5 MDIC modules, hereafter referred to as $EM$. The core idea of MDIC is to divide all the feature maps generated after the previous convolutional layer into four equal parts, named $[EM1, EM2, EM3, EM4]$, and implement different flip angles for each local feature map, named $[FEM1, FEM2, FEM3, FEM4]$. Then, four deeper feature maps are generated by convolution for each local feature map after flipping. After all convolutions, we first introduce the non-linear ReLU activation function layer and then introduce the batch normalization (BN) layer [40]. The reason for this order is that when a feature enters the ReLU layer first, a feature with few semantics or a high amount of redundant information is converted from a negative value to 0, thereby improving computational efficiency. Subsequently, the feature map after convolution is flipped opposite of the original flip method and then transformed into $[EM1, EM2, EM3, EM4]$. Through the MDIC model, the semantic information and multi-scale semantic information of different flipped feature maps are fully obtained, which greatly strengthens the local feature maps of medical images. At the same time, to obtain the semantic information from the global feature map, we introduce a traditional convolution module with a different scale from that of the MDIC module. The purpose of this module is to obtain the global feature map of the medical image. By fully mining local features and global features, medical images are provided with more semantic information, and even through a small-sample data set or a small number of fitters, more accurate prediction results can be obtained. The main formula for the process of MDIC is as follows:

$$(EM_1, EM_2, EM_3, EM_4) = Average(EM) \quad (1)$$

where $(EM_1, EM_2, EM_3, EM_4)$ represents the four local feature maps in the MDIC module; $Average$ represents the function of dividing the number of feature maps equally.

The formula for flipping local feature maps is as follows:

$$REM = \begin{cases} REM_1, X \\ REM_2, P(X,(0,1,3,2)) \\ REM_3, F(X,[2,3]) \\ REM_4, F(F(X,[2,3]),[2]) ) \end{cases} \quad (2)$$

where $REM$ represents the new local feature maps generated after the local feature maps are flipped; $P$ represents the third and fourth dimensions of the local feature maps in the DT-Net algorithm that are flipped by 90 degrees; $F$ represents the features of the third and fourth dimensions of the local feature maps in the DT-Net algorithm that simultaneously exchange positions in the local feature maps; $X$ represents the input feature maps.

We describe the process of feature map flipping in detail through Figure 3 below.

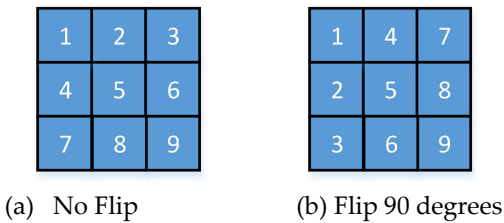

(a)  No Flip                    (b)  Flip 90 degrees

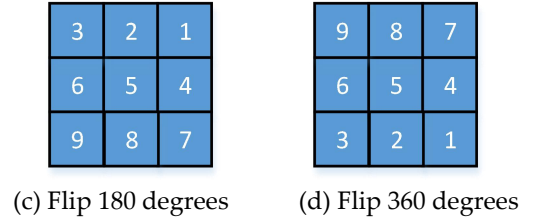

(c) Flip 180 degrees           (d) Flip 360 degrees

Fig. 3. The results of flipping at different degrees.

After the global feature map passes the above processing steps, we obtain a multi-directional, multi-scale local feature map. Through the enhancement operation on the local feature map, even with a small number of fitters, a higher quality feature map can be obtained. Next, for each multi-scale local feature map, the compression and integration of features in each local feature map is realized through the 1*1 convolution kernel so that the feature points in each feature map can obtain more semantic information while reducing the amount of required parameter calculations. Then, the integrated feature map is filtered through "threshold convolution". The main purpose of this step is to eliminate a large number of redundant features while retaining more valuable feature points. The specific flow of the algorithm is given in detail in the "threshold convolution" section. Finally, the feature map is integrated and condensed through the 1*1 integration network again, and the translation invariance of the feature map is retained to the greatest extent through maximum pooling.

### 3.3 Decoder: MDIC Module

We abbreviate the MDIC of the decoder as DM. The main differences between the MDIC modules in the encoder and in the decoder are as follows: 1. The DM introduces the EM feature map, so the EM feature map serves as the prior semantic information of the DM feature map in the upsampling process. To avoid increasing the amount of parameter calculations, we use a residual computer to realize the semantic fusion of the feature map. 2. We do not introduce threshold convolution in the DM, because its purpose is to obtain more semantic information during the upsampling process, thereby obtaining higher quality feature maps. The specific model structure is shown in the "Decoder: MDIC module" in Figure 2. In the DM, all our upsampling layers use bilinear interpolation to extend the feature map from a high-dimensional space to a low-dimensional space. Figure 4 shows the detailed structure of the DM module.

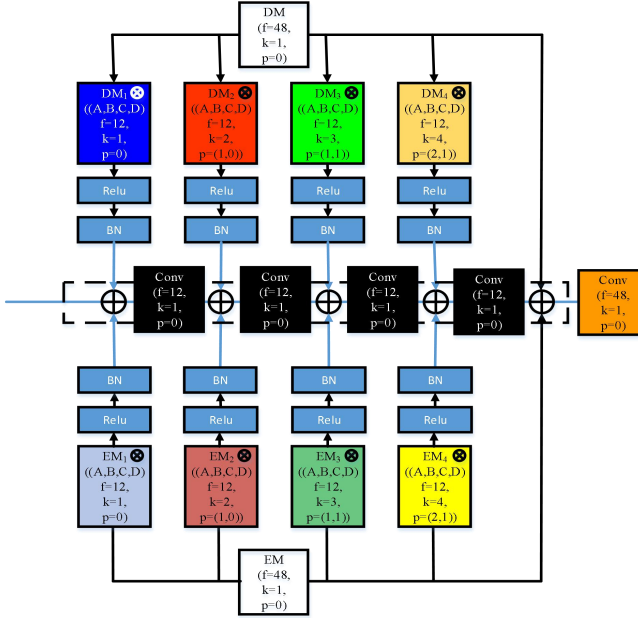

Fig. 4. Illustration of the fourth module of the DM. "f" represents the number of input fitters; "k" represents the size of the convolution kernel; "p" represents the zero padding after the convolution operation; ⊕ represents the residual calculation of the EM module and the DM module; and ⊗ represents the splicing operation of all local and global feature maps.

It can be seen from Figure 4 that the global feature maps of the EM and DM are divided into four equal local feature maps (A, B, C, D in the figure represent four groups of local feature maps with different scales in the same direction). The new local feature map merges the original feature maps that are in the same direction and have different convolution kernel sizes to obtain richer semantic information about the medical image. To retain the original feature information when mining new features, we use one convolution kernel operation. To further improve the quality of the feature map, we make a residual computer for each local feature map of the EM and DM and obtain richer semantic information. Finally, all the local feature maps and global feature maps are merged to obtain richer semantic information about medical images.

### 3.4 Threshold Convolution

Some of the feature points contain semantic information but also a lot of redundant information. If the amount of redundant information is too large, other feature points may be affected, which ultimately reduces the effectiveness of medical image segmentation. To better retain semantically rich features and reduce redundant features, we construct a threshold after the convolution operation, which we call threshold convolution. The idea of threshold convolution is to use the ReLU activation function while retaining the features of the feature map output by the previous layer (because the weight value generated by ReLU is between (0, 1) and contains more semantic information features, the greater the point weight value is, the smaller the weight value as the semantic information decreases.) Weight values are generated for all feature points. At the same time, in the threshold convolution module, we build 2 sets of 0 and 1 array with the same number of feature maps. The purpose of this is to convert the feature point into 0 when the weight value generated by the ReLU activation function is less than the set threshold value, thereby reducing the influence of redundant feature points on medical image segmentation tasks. Finally, the transfer invariance of semantically rich feature points is realized by dot multiplication, as shown in subgraph (a) of Figure 5. The judgment of as the weight value by ReLU leads to the loss of some feature points in the feature map. A large number of iterations through the DT-Net algorithm affects the final medical image segmentation task. Therefore, we make further improvements to the subgraph in Figure 5 (a), adding a very small positive value (1-e10) to the feature points that are equal to 0 in the final feature map. This is done to reactivate the iterative parts of feature points to obtain more semantic information.

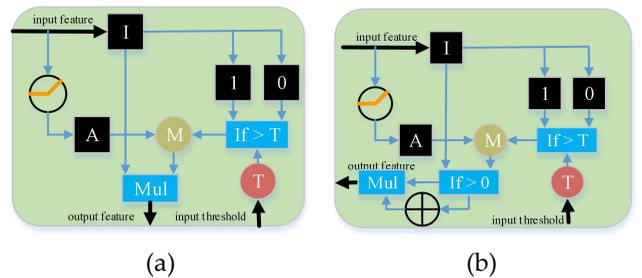

Fig. 5. Two methods of threshold convolution. "I" represents the input feature maps; "A" represents the weight value generated by the ReLU activation function; ☺ represents the activation function; "M" represents the threshold value compared with the weight values of the feature maps; "T" represents the input threshold; "1" represents the feature maps with the same number of 1s as the feature maps; "0" represents the feature maps with the same number of 0s as the feature maps; "Mul" represents the dot multiplication operation; ⊕ represents the addition operation; the black arrows represent the inputs and outputs of the feature maps; and the blue arrows represent various operations in the threshold convolution strategy.

# 4 EXPERIMENTS AND RESULTS

## 4.1 Datasets

In this paper, we select 2 public medical image data sets for multi-segmentation tasks to verify the DT-Net model.

**CHAOS data set** [41]. The goal of this multi-segmentation medical image data set is to segment human abdominal organs (for example, liver, left kidney, right kidney, spleen). Our model performs a segmentation prediction on the "T1-DUAL in phase" part of the data set because this part of the MRI image has an accurate ground truth. We scramble the overall data before training and then distribute it. Finally, we divide the selected 647 MRI images into a training set and a test set with a 3:2 ratio for the number of images. The image size of this data set is 256*256.

**BraTS data set** [42], [43]. This data set is a multi-class brain segmentation data set. Since the validation set of this data set does not have a ground truth, we only obtain training samples from 210 higher-grade glioblastoma (HGG) and 75 lower-grade glioma (LGG) patients in the training set. To further verify the performance of our proposed DT-Net algorithm on a small-sample data set, we select 2 sets of images from each patient sample (each sample has 155 sets of images). Each set of images consists of 4 MRI image compositions (including FLAIR, T1, T1CE and T2). We label all pixels in accordance with the requirements of the data set: oedema (ground truth of 1), non-enhancing solid core (ground truth of 2), enhancing core (ground truth of 3), and remaining images with ground truths of 0. Finally, to reduce the workload, we fill the original images with scales of 240*240 to 256*256 by zero padding.

## 4.2 Experimental Evaluation Criteria

The CHAOS data set is composed of four foreground classes, Liver, Kidney R, Kidney L and Spleen, and one background class. We evaluate the segmentation results of each class through the correctness rate for each class. At the same time, we also use sensitivity (Sen), specificity (Spec) and Dice scores (Dice) to further evaluate the performance of the model. The formula for each evaluation index is as follows:

$$Accuracy = (TP+TN) / (TP+FN+FP+TN) \quad (3)$$

$$Sen = TP / (TP + FN) \quad (4)$$

$$Spec = TN / (TN + FP) \quad (5)$$

$$Dice = (2*TN) / (2*TN + FN + FP) \quad (6)$$

where $TP$ represents that the sample itself is the foreground class, and the prediction is correct; $TN$ represents that the sample itself is the foreground class, but the prediction is wrong; $FP$ represents that the sample itself is the background class, but the prediction is the foreground class; $FN$ represents that the sample itself is the background class, and at the same time, the prediction is correct.

WT (including all four tumour structures), ET (including all tumour structures except "oedema"), and TC (only containing the "enhancing core" structures that are unique to high-grade cases) in the BraTS data set are used separately. Three groups of evaluation indicators, $Dice^+$ $Sens^+$ and $Spec^+$, are used. The specific formulas are as follows:

$$Dice(M,N)^+ = |M_1 \wedge N_1| / ((|M_1| + |N_1|) / 2) \quad (7)$$

$$Sens(M,N)^+ = |M_1 \wedge N_1| / |N_1| \quad (8)$$

$$Spec(M,N)^+ = |M_0 \wedge N_0| / |N_0| \quad (9)$$

where $M_0$ and $N_0$ are pixels with $M = 0$ and $N = 0$, respectively. The operator $\wedge$ is the logical AND operator, $| \ |$ is the size of the set (i.e., the number of pixels belonging to it), and $M_1$ and $N_1$ are the sets of pixels with $M = 1$ and $N = 1$. The value of 1 represents all the current evaluation objects, and 0 represents the remaining objects.

## 4.3 Setting of the Experimental Parameters

This experiment is performed using the PyTorch library. The server contains an NVIDIA Tesla V100 16 GB graphics card. All experiments use the Adam optimizer: the learning rate is set to 0.001; the attenuation of the first-order matrix exponent is set to 0.9; the attenuation of the second-order matrix exponent is set to 0.999; other parameters are set to the default values. All experiments use the CrossEntropyLoss loss function (which includes the loss function and the Softmax activation function). During the training process, all training models are iterated 300 times.

## 4.4 Model Performance

In this section, we compare some classic medical image segmentation algorithms and some recent segmentation algorithms that have achieved good results on segmentation tasks. The CHAOS and BraTS data sets are not subject to any other preprocessing aside from disrupting the order of the data sets. The experimental results based on the CHAOS data set are shown in Table 1, and the visualization results are shown in Figure 6; the experimental results based on the BraTS data set are shown in Table 2, and the visualization results are shown in Figure 7.

TABLE I SEGMENTATION RESULTS FOR THE CHAOS DATA SET OBTAINED WITH EACH MODEL

| Data set | Model | Liver | Kidney L | Kidney R | Spleen | Sen | Spec | Dice |
|----------|-------|-------|----------|----------|--------|-----|------|------|
| CHAOS | U-Net [31] | 91.23 | 79.01 | 79.66 | 83.35 | 89.69 | 88.24 | 99.18 |
| | AttU-Net [46] | 90.37 | 77.98 | 78.68 | 84.16 | 89.79 | 87.46 | 99.19 |
| | R2AttU-Net [47] | 87.79 | 43.86 | 57.08 | 69.74 | 83.39 | 78.98 | 98.81 |
| | Seg-Net [45] | 89.43 | 78.48 | 78.42 | 78.94 | 87.97 | 86.20 | 99.06 |
| | PSP-Net [44] | 89.94 | 74.68 | 70.02 | 79.54 | 87.83 | 85.50 | 99.03 |
| | CE-Net [20] | 93.03 | 81.14 | 79.52 | 86.71 | 90.08 | 89.98 | 99.21 |
| | DenseASPP [49] | 91.21 | 72.85 | 78.34 | 82.23 | 88.17 | 87.34 | 99.05 |
| | MS-Dual [50] | 91.87 | 80.58 | **80.59** | 84.03 | 89.38 | 88.82 | 99.15 |
| | Deeplabv3 [51] | 91.03 | 72.98 | 76.44 | 81.46 | 88.55 | 86.96 | 99.09 |
| | DT-Net | **93.55** | **81.34** | 80.34 | 88.92 | **90.32** | **90.55** | **99.25** |

TABLE II SEGMENTATION RESULTS FOR THE BraTS 2018 DATA SET OBTAINED WITH EACH MODEL

| Data set | Model | Dice⁺ | | | Sens⁺ | | | Spec⁺ | | |
|----------|-------|-----|-----|-----|-----|-----|-----|-----|-----|-----|
| | | ET | WT | TC | ET | WT | TC | ET | WT | TC |
| BraTS | U-Net [31] | 69.41 | 68.73 | 67.11 | 75.58 | 69.23 | 78.81 | 98.99 | 99.52 | 98.53 |
| | AttU-Net [46] | 63.72 | 64.79 | 63.39 | 63.57 | 63.10 | 61.27 | 98.99 | 99.44 | 98.36 |
| | R2AttU-Net [47] | 59.43 | 56.38 | 61.08 | 56.46 | 50.64 | 49.33 | 98.97 | 99.71 | 98.25 |
| | Seg-Net [45] | 48.73 | 49.32 | 45.18 | 56.98 | 53.36 | 45.12 | 98.02 | 98.66 | 97.36 |
| | PSP-Net [44] | 58.36 | 61.59 | 56.29 | 57.26 | 59.79 | 63.33 | 99.03 | 99.44 | 98.22 |
| | CE-Net [20] | 79.08 | 79.40 | 78.31 | **81.88** | **80.29** | 82.23 | 99.36 | 99.64 | 99.04 |
| | DenseASPP [49] | 66.73 | 68.69 | 63.84 | 67.18 | 68.48 | 66.22 | 99.18 | 99.52 | 98.60 |
| | MS-Dual [50] | 79.25 | 79.30 | 78.28 | 77.66 | 77.34 | 81.82 | 99.51 | 99.78 | 99.07 |
| | Deeplabv3 [51] | 71.22 | 72.32 | 68.30 | 68.29 | 69.81 | 66.96 | 99.40 | 99.72 | 98.84 |
| | DT-Net | **80.99** | **80.65** | **81.36** | 81.01 | 78.88 | **86.76** | **99.52** | **99.82** | **99.13** |

Table I shows the experimental results of each model on the CHAOS data set. From the overall data, it can be seen that the prediction accuracies of Kidney L and Kidney R is lower than those of the other two classes because the total number of samples in other classes is greater than the total number of samples in these two classes. Although the MS-Dual algorithm's recognition of Kidney R is 0.25% higher than that of our proposed model (DT-Net), the DT-Net algorithm performs 1.68%, 0.76%, and 4.89% better than the MS-Dual algorithm in the prediction of the other three categories, Liver and Kidney L, Spleen, respectively. The result of the DT-Net algorithm is 0.24% higher, the Spec value is 0.57% higher, and the Dice value is 0.03% higher than those of the CE-Net with the second-highest overall prediction results, which proves that our proposed model performs well on the CHAOS data set. The results on the CHAOS data set show that the DT-Net algorithm proposed in this article can obtain feature points containing more semantic information through local feature map enhancement and global feature fusion even with a small number of samples of multi-classification tasks, which proves that the algorithm has a strong feature learning ability and a strong feature fusion ability. Figure 6 shows the visualization results of each model om the CHAOS data set.

Table II shows the prediction results of each model on the BraTS data set. This data set is a multi-modal data set. The input is composed of four different types of MRI medical images: FLAIR, T1, T1CE and T2. Judging from the experimental results of each model, our proposed DT-Net is superior to other models based on the overall experimental results in terms of Dice⁺ and Spec⁺. At the same time, the prediction result for "enhancing core" is 4.54% higher than that of the CE-Net. For the various evaluation indicators, we find that the CE-Net and MS-Dual obtain good prediction results. The reason for this is that two modules are proposed in the CE-Net network for obtaining more semantic information: 1. The principle of DAC encoding to generate high-level semantic feature mapping is used to obtain multiple receptive fields for extracting the features of targets of various sizes; 2. The RMP module is used to obtain the contextual semantic information encoded by receptive fields of different sizes through pooling strategies of different scales. The principle is that pooling can yield the invariance of feature transfers. In the MS-Dual model, a large number of global features are first obtained through a multi-scale strategy, and then the features are integrated into the two attention mechanism modules, the channel attention module and the position attention module, to achieve the fusion of local and global features and eliminate many redundant features. Other models possess certain feature-learning capabilities for the BraTS data set. Among them, U-Net, DenseASPP, and Deeplabv3 have good predictive capabilities. This is because that these models introduce the prior features of the encoder in the decoding process, making the decoder generate a feature map with more prior semantic information. Simultaneously, the experimental results of AttU-Net and R2AttU-Net are worse than those of U-Net. We have explained that the traditional attention mechanism is difficult to apply to complex medical images. The DT-Net mentioned in this article obtains higher-quality deep feature maps by strengthening local feature maps and fusing global feature maps. We can obtain higher-quality feature maps with only a small number of fitters (in Part E of this chapter, we analyse the number of fitters for each model and the total amount of final calculation parameters in detail). To further improve the quality of the feature map and eliminate a host of redundant features, we propose a threshold convolution in the encoder. The results in Table 2 show the powerful performance of the DT-Net model and further verify the feature-learning ability of the algorithm. Figure 7 shows the visualization results of each model on the BraTS data set.

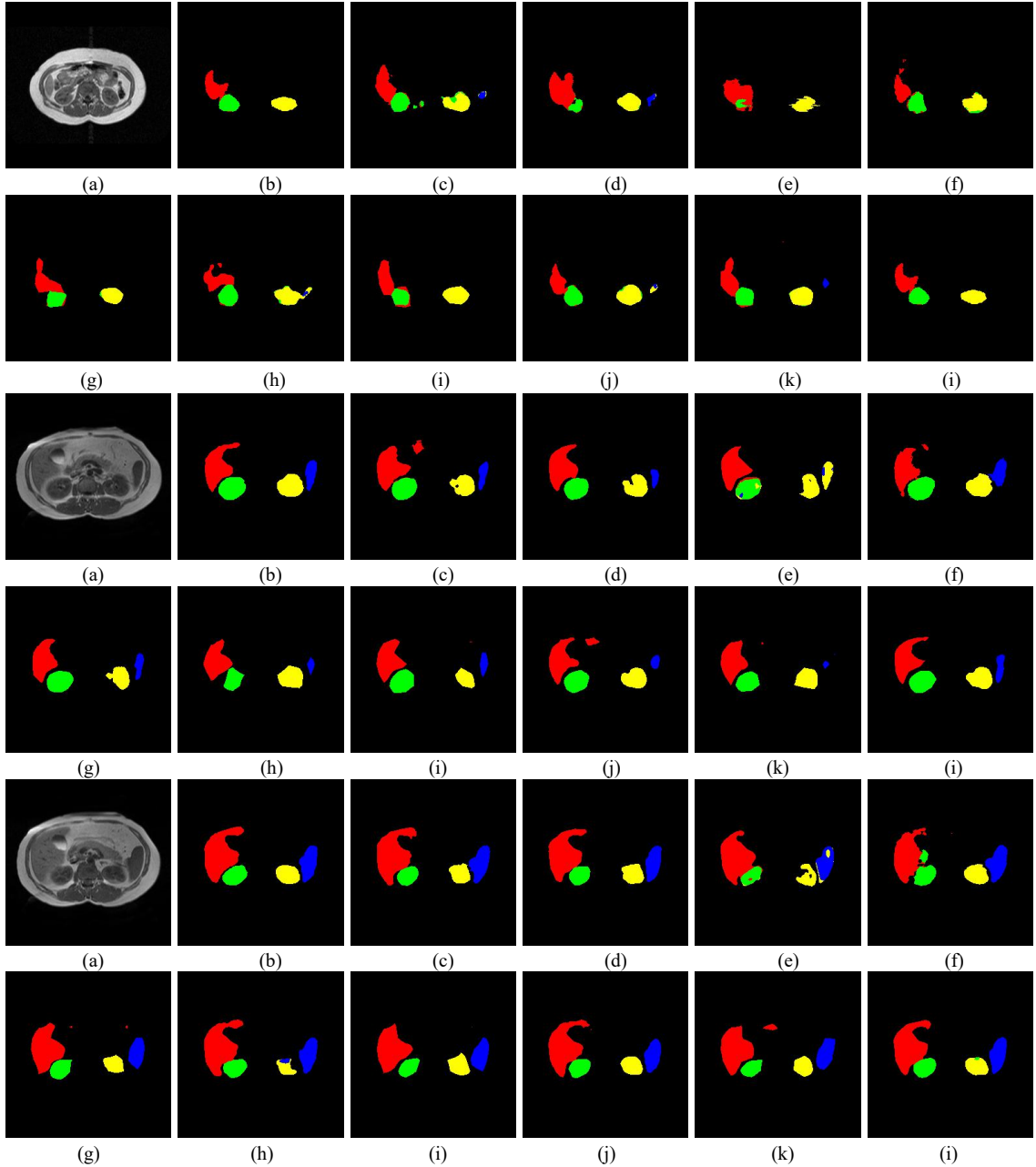

Fig. 6. Segmentation results of each model for the CHAOS data set. (a) Original image. (b) Ground truth. (c) U-Net. (d) AttU-Net. (e) R2AttU-Net. (f) Seg-Net. (g) PSP-Net. (h) CE-Net. (i) DenseASPP. (j) MS-Dual. (k) Deeplabv3. (i) DT-Net. The colours indicate Liver (red), Kidney L (green), Kidney R(yellow), and Spleen (blue).

Figure 6 shows the visualization results for the 3 sets of images of the CHAOS data set. It can be seen from the first group that the DT-Net algorithm proposed in this article obtains the most accurate prediction results for Liver, Kidney L, and Kidney R. From the segmentation results, we see that other models produce much noise or lose some key features, which causes confusion in the results. In the second set of experimental visualizations, the DT-Net algorithm is very effective at edge segmentation, and the generated contours of each organ classification are relatively smooth. At the same time, compared with other models, it generates less noise for the Liver class and has a stronger predictive ability for the Spleen class (the sample size of Spleen in the reorganized data set is less than those of the other classes). From the third set of segmented imaging results, it can be clearly seen that the DT-Net algorithm and the MS-Dual algorithm have powerful capabilities for learning the details of medical images. From the above three sets of visualization results, it can be seen that the proposed model (DT-Net) has strong capabilities for denoising, edge recognition, and detail processing through local feature enhancement and global feature fusion.

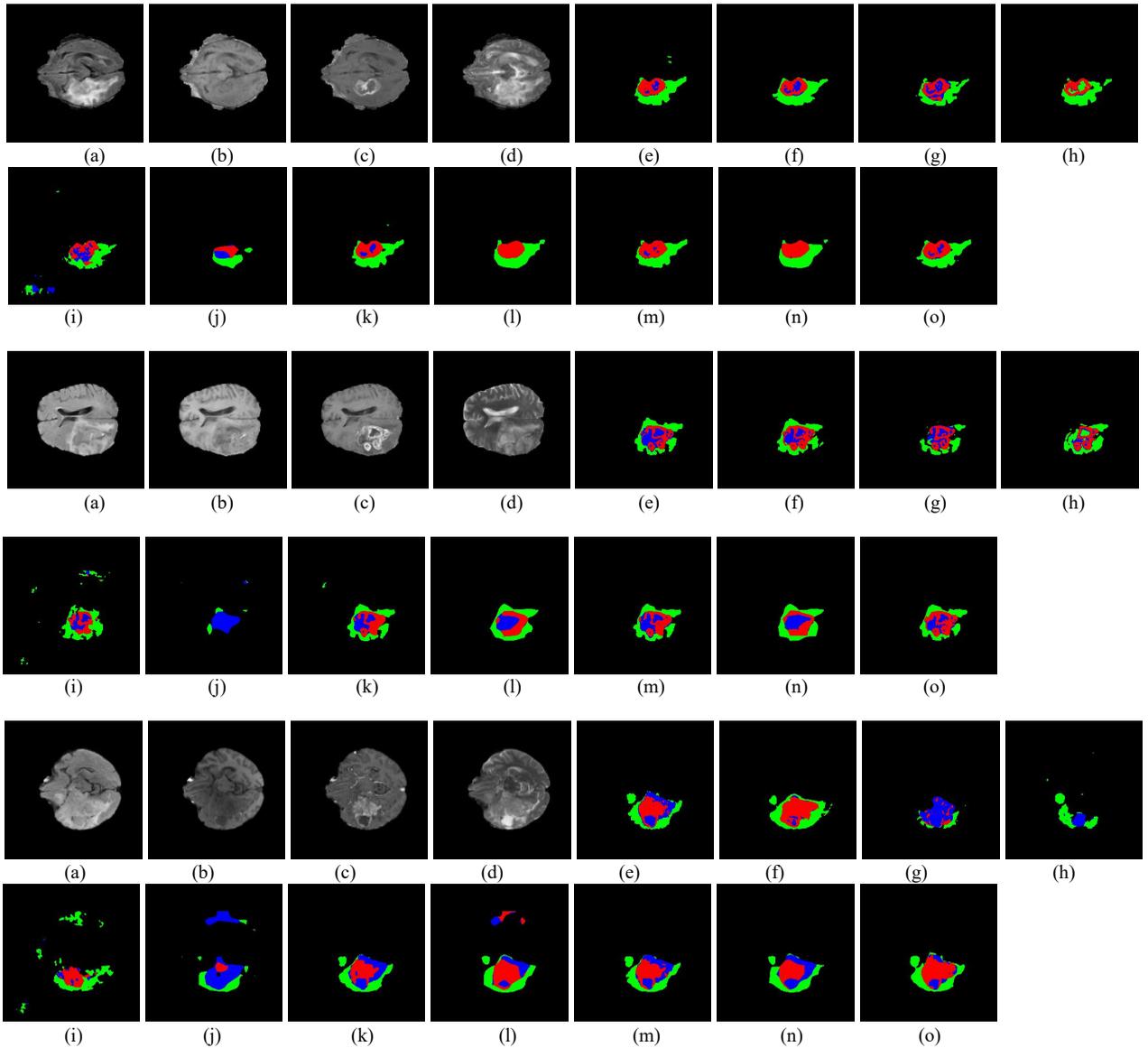

Fig. 7. Segmentation results of each model for the BraTS 2018 data set. (a) Flair. (b) T1. (c) T1ce. (d) T2. (e) Ground truth. (f) U-Net. (g) AttU-Net. (h) R2AttU-Net. (i) Seg-Net. (j) PSP-Net. (k) CE-Net. (l) DenseASPP. (m) MS-Dual. (n) Deeplabv3. (o) DT-Net. The colours indicate oedema (green), non-enhancing solid core (red), and enhancing core (blue).

Figure 7 shows the results of the segmentation experiment on the BraTS dataset. From the first set of segmentation results, it can be seen that the CE-Net model achieves better performance for the oedema class, but DT-Net achieves better results for the non-enhancing solid core class and the enhancing core class. In particular, DT-Net produces the best edge processing effect. In the second set of data, R2AttU-Net learns the oedema class and the enhancing solid core class very well, but its effect on the enhancing core class is very poor. PSP-Net can only learn the enhancing core class well. Seg-Net and CE-

Net generate small amounts of noise. DenseASPP and Deeplabv3 also lack small sample texture for multi-modal data. MS-Dual and the DT-Net algorithm proposed in this article yield better segmentation results for multi-modal data sets. From the third set of experiments, we can clearly see that for DT-Net,, the prediction results of each class and the edges between classes are better than those of the other algorithms. Finally, it is verified that the DT-Net model mentioned in this article plays an active role in feature mining and learning.

### 4.5 The Number of Fitters and Calculation Parameters of Each Model

Convolution operations can generate various feature maps due to different weights. Usually, the number of feature maps is equivalent to the number of fitters. More feature maps produce more semantic information and generate a large number of calculation parameters. For this reason, we enhance the local feature map through the principles of multiple directions and multiple scales without increasing the number of global feature maps. Through Table 3 below, we count the number of feature maps and operation parameters of each model in Part D of this chapter and directly analyse the operation efficiency of each algorithm.

TABLE III NUMBER OF FITTERS AND PARAMETERS OF EACH MODEL

| Model | | |
|---|---|---|
| | Fitters | Total Parameters |
| U-Net [31] | [64, 128, 256, 512, 1024] | 31,042,629 |
| AttU-Net [46] | [64, 128, 256, 512, 1024] | - |
| R2AttU-Net [47] | [64, 128, 256, 512, 1024] | - |
| Seg-Net [45] | [64, 128, 256, 512, 512] | 29,444,741 |
| PSP-Net [44] | [64, 128, 256, 512, 1024] | 46,578,117 |
| CE-Net [20] | [32, 64, 128, 256, 512, 2048] | 38,964,060 |
| DenseASPP [49] | [32, 64, 128, 256, 512, 1024] | 46,161,989 |
| MS-Dual [50] | [64, 128, 256, 512, 1024] | 100,763,834 |
| Deeplabv3 [51] | [64, 128, 256, 512, 1024, 2048] | 39,039,685 |
| DT-Net | [24, 48, 96, 192, 192] | **5,272,277** |

Table III shows the number of fitters used in each comparison model and the total parameters for each model (on the CHAOS data set). According to the experimental results in Part D of this chapter, it can be clearly observed that the CE-Net algorithm and the MS-Dual algorithm both generate 2048 deep feature maps, thereby obtaining more semantic information and producing good results for the two data sets in this article. Since the CE-Net algorithm uses hole convolution, the computational complexity of the model is reduced. The MS-Dual algorithm uses traditional convolution, so the total amount of parameters is the largest among all models. Although the total amount of parameters for AttU-Net and R2AttU-Net is not printed, these algorithms are based on U-Net with added attention, and so the total amount of parameters is definitely greater than that of U-Net. Our proposed model, DT-Net, uses only 24 fitters in the first layer and then doubles the number of fitters in each successive sub-module (the fifth sub-module does include a doubling step). In the end, the total number of parameters of the DT-Net algorithm is only 5272277. In the two sets of multi-classified small-sample data, state-of-the-art segmentation results are obtained, which fully proves the superior performance of the algorithm proposed in this paper.

### 4.6 Feature Map Visualization of MDIC

Figure 8 shows the 4 sets of MDIC encoder modules. "A" represents the first layer of the MDIC encoder (the size of the feature map is 128*128); "B" represents the second layer of the MDIC encoder (the size of the feature map is 64*64); "C" represents the third layer of the MDIC encoder (the size of the feature map is 32*32); and "D" represents the fourth layer of the MDIC encoder (the size of the feature map is 16*16). It can be seen from the feature maps of each layer that even with convolutions of the same scale, feature points containing different semantic information can also be obtained by flipping the feature map. As the scale of the feature map shrinks, the difference between the feature maps generated by different flips becomes significantly larger.

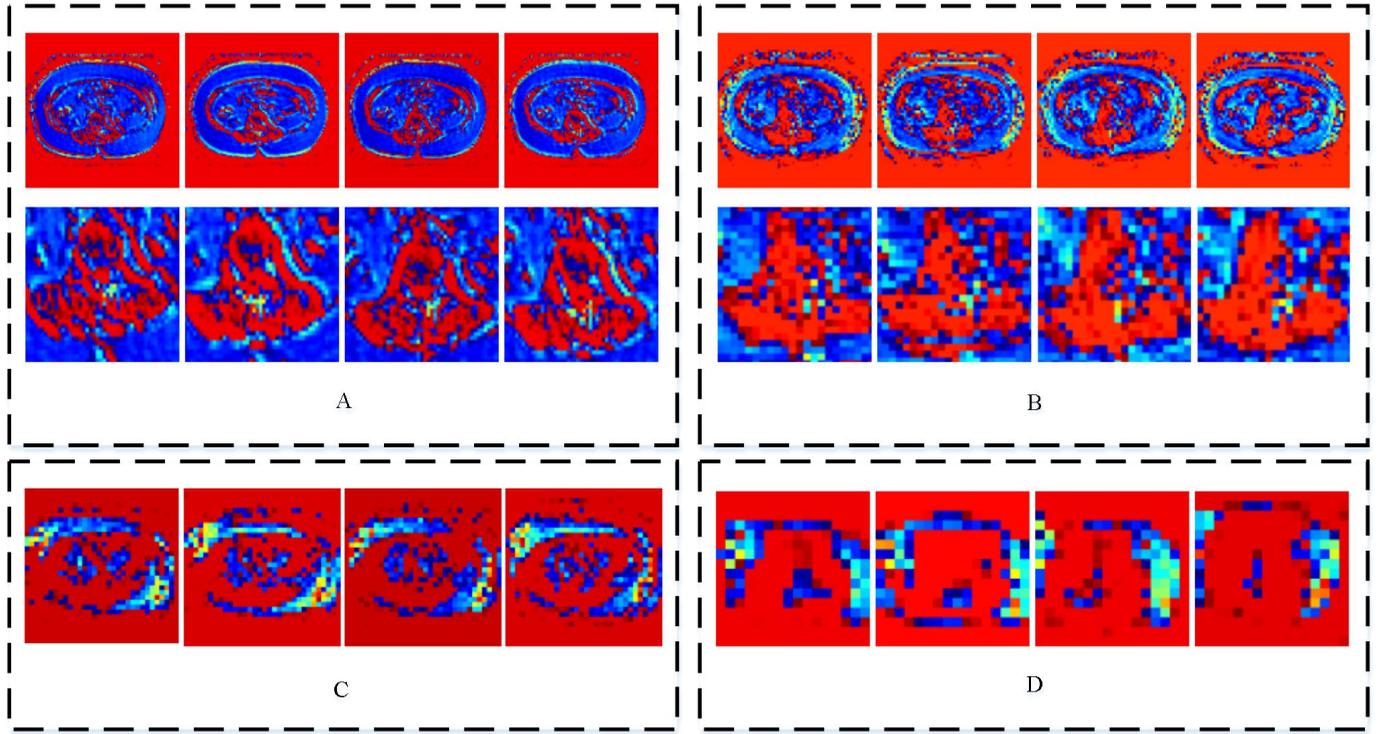

Fig. 8. Feature Map Visualization of MDIC. The four modules, A, B, C, and D, are the local feature maps of the first part of the MDIC encoder, and the feature maps of each module from left to right are flipped 0 degrees, 90 degrees, 180 degrees, and 360 degrees after passing the same convolution kernel (the convolution kernel size of the feature map in Figure 8 is equal to 3). A new feature map is generated by scaling.

## 4.7 The Influence of Different Thresholds on The Experiment

Some feature points in the feature map generated by traditional convolution contain a large number of redundant features, which often reduces the prediction accuracy. To further solve this problem, we set a threshold to retain the non-redundant features, thereby improving the quality of the feature map. We use line graphs to specifically analyse the importance of the threshold convolution mentioned in this article. The line diagrams are shown in Figure 9 below.

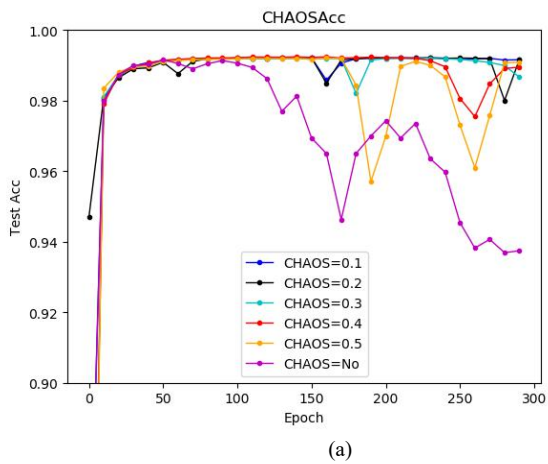

(a)

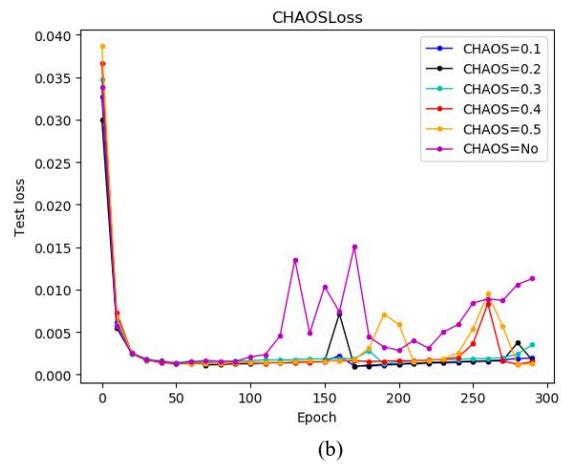

(b)

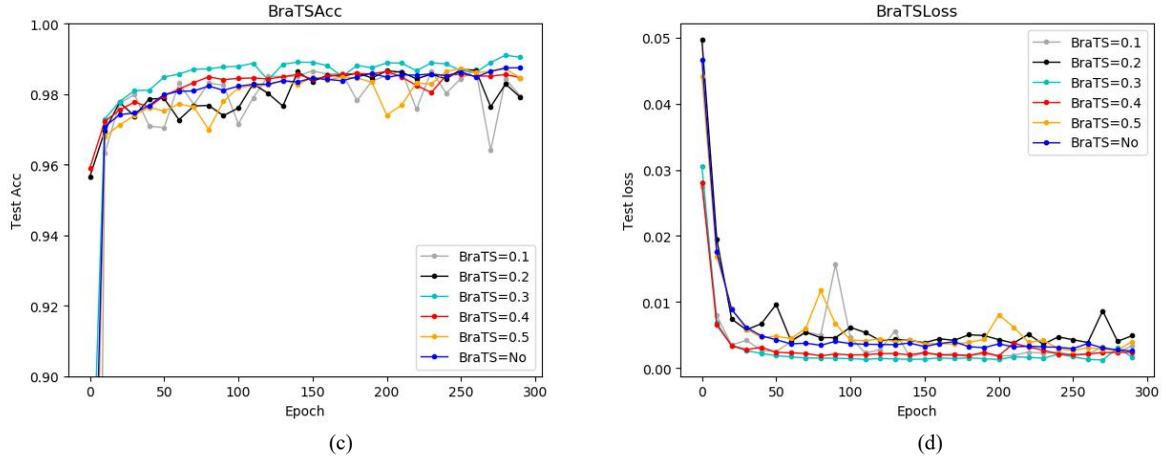

Fig. 9. (a) Accuracy results with different thresholds for the CHAOS data set. (b) Loss results with different thresholds for the CHAOS data set. (c) Accuracy results with different thresholds for the BraTS data set. (d) Loss results with different thresholds for the BraTS data set.

The 4 line graphs in Figure 9 represent the testing process of DT-Net for the CHAOS data set and the BraTS data set.

(a) This line graph shows the process of testing the accuracy on the CHAOS data set. When the threshold is set to 0.1, the training process is the most stable. As the threshold is increased, a large number of features are lost, resulting in unstable testing. When the threshold is not set, the test process has an overfitting phenomenon, so we infer that the threshold weight for most features of the CHAOS data set is approximately 0.1.

(b) This line graph shows the process of testing the loss value on the CHAOS data set. If the threshold convolution is not set, overfitting also occurs during the test. (c) and (d) These two line graphs represent the testing of accuracy and loss on the BraTS data set. It can be seen intuitively from the line graph that when the threshold is set to 0.3, the best prediction results for both accuracy and loss are obtained. The results for thresholds of 0.2 and 0.4 are slightly smaller than those for 0.1 and 0.5. The results for the BraTS data set are worse with thresholds of 0.3 and 0.4 than they are when a threshold is not set at all. There is no overfitting phenomenon in the BraTS data set. We speculate that the generated deep features are more robust due to the acquisition of semantic information of multiple modalities. From this subsection, we verify the feasibility and effectiveness of threshold convolution.

(c)

## 4.8 The Impact of Different Strategies on DT-Net

Different models obtain different feature maps through differently constructed structures, thereby describing different semantic information. To further verify the rationality of the structure of the algorithm proposed in this article, we verify all the structures in the model one by one. To highlight the results of each strategy more intuitively, we use the polar axis pie chart to describe the comparative experimental results of various strategies. "1" represents the MDIC strategy; "2" represents the threshold convolution strategy; "3" represents the introduction of a priori feature maps in the decoder. "*" means that an extra decimal is not added after the threshold division. Table IV shows the parameters and thresholds of the different strategies (the left side is the number of total parameters for the CHAOS data set, and the right side is the number of total parameters for the BraTS data set). Figure 10 shows the experimental results of different strategies on the CHAOS data set; Figure 11 shows the experimental results of different strategies on the BraTS data set.

TABLE IV THE PARAMETERS AND THRESHOLDS FOR DIFFERENT STRATEGIES.

| Model | Total Parameters | Threshold |
|---|---|---|
| DT-Net-* | 5272277,5275852 | 0.1 |
| DT-Net-no-1 | 7651541,7655116 | 0.1 |
| DT-Net-no-2 | 5272277,5275852 | - |
| DT-Net-no-3 | 5272277,5275852 | 0.1 |
| DT-Net-no-1-2 | 7651541,7655116 | - |
| DT-Net | 5272277,5275852 | 0.1 |

Figure 10 clearly shows that DT-Net obtains the best experimental results in all four categories. In the classification of Liver, the "3" strategy has the greatest impact; for Kidney L and Kidney R, the combined effect of the "1" strategy and the "2" strategy has the greatest impact on the data, which proves that MDIC and threshold convolution play an active role in small samples; for Spleen, the "2" strategy or the combination of the "1" strategy and "2" strategy has the greatest impact. Finally, it can be analysed from the above figure that each strategy with DT-Net has a positive effect on the CHAOS data set.

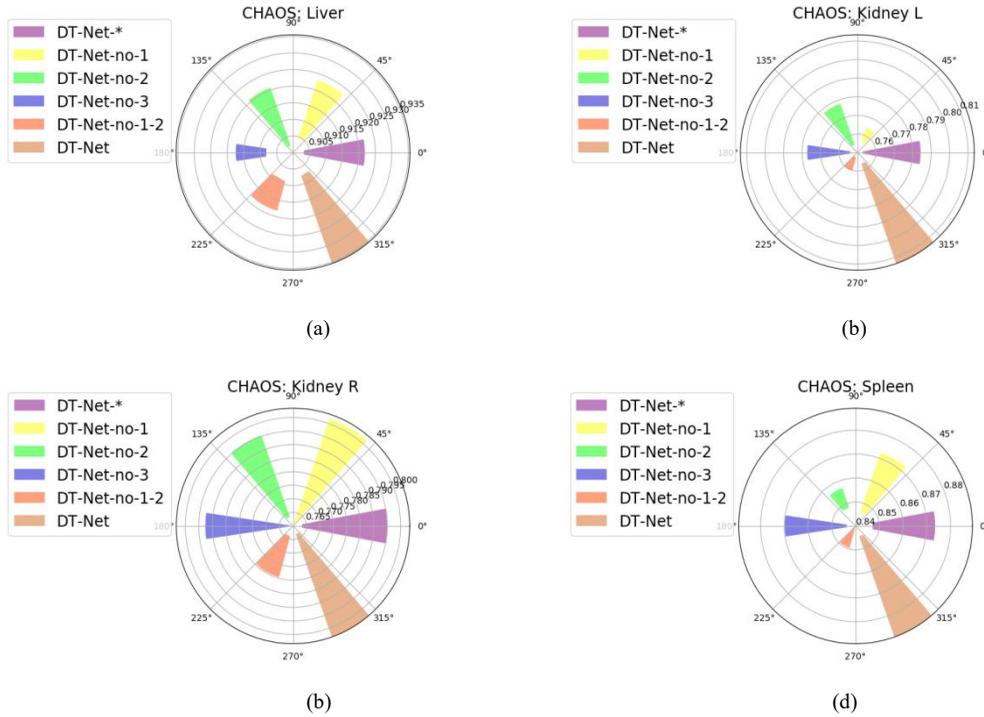

Fig. 10. The experimental results of different strategies for the CHAOS data set are shown in the polar axis pie chart. (a) Liver category. (b) Kidney L category. (c) Kidney R category. (d) Spleen category.

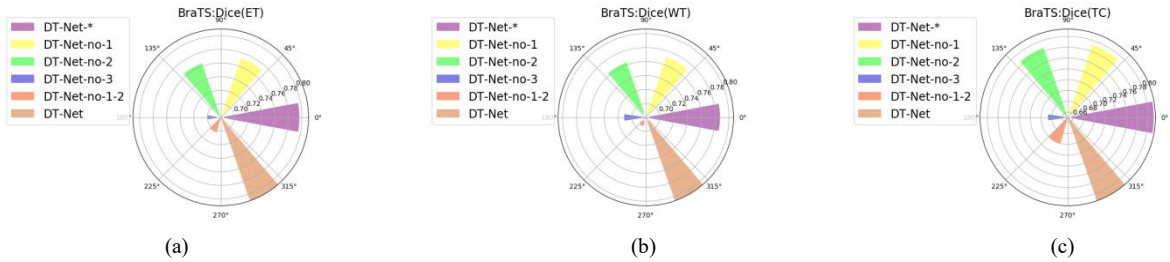

Fig. 11. The experimental results of different strategies for the BraTS data set are shown in the polar axis pie chart. (a) Oedema (ET). (b) Non-enhancing solid core (WT). (c) Enhancing core (TC).

In the BraTS data set, through ET, we can clearly see that the combination of the "1" strategy and the "2" strategy has the greatest impact; WT's experimental results reflect that the "3" strategy has more impact when it does not predict "non-enhancing"; the result of TC also suggests that the combination of the "1" strategy and "2" strategy has the greatest impact. After the threshold is divided, 0 is converted into an extra decimal to obtain a better classification result(The effects of CHAOS data set and BraTS data set are both obvious). The BraTS data set further verifies the feasibility of our proposed strategy.

## 5 CONCLUSIONS

According to the experimental results verified from multiple directions on the above two multi-classification data sets, it is feasible to use the enhancement of local feature maps and the introduction of three groups of prior feature maps. These strategies can be implemented with small-sample data sets. With more semantic medical image information, threshold convolution, can also play an important role in eliminating redundant features. In the task of image segmentation, most feature fusion is done via the introduction of the encoder's prior feature map through the decoder. There are often few effective fusion strategies in the encoder module. For this reason, in the next study on medical image segmentation, we will construct an effective fusion strategy and realize feature fusion in the encoder module to provide more prior semantic information for the final image segmentation.

## ACKNOWLEDGMENT

The authors wish to thank A, B, C. This work was supported in part by a grant from XYZ.